\documentclass{icorsdssv2024}
\usepackage{float} 
\usepackage{graphicx}
\usepackage{booktabs}
\usepackage{adjustbox}
\usepackage{multirow}
\usepackage{amsmath}
\usepackage{caption}
\usepackage{subcaption}

\begin{document}

\mytitle{Earthquake Damage Grades Prediction using An Ensemble Approach Integrating Advanced Machine and Deep Learning Models}
\mytitlerunning{3rd International Conference on Applied Mathematics in Science and Engineering}

\myauthor{Anurag Panda, Gaurav Kumar Yadav}

\myaddress{Department of Computer Science and Engineering, Institute of Technical Education and Research, Siksha 'O' Anusandhan University, Bhubaneswar, Odisha}

\abstract{
In the aftermath of major earthquakes, evaluating structural and infrastructural damage is vital for coordinating post-disaster response efforts. This includes assessing damage's extent and spatial distribution to prioritize rescue operations and resource allocation. Accurately estimating damage grades to buildings post-earthquake is paramount for effective response and recovery, given the significant impact on lives and properties, underscoring the urgency of streamlining relief fund allocation processes. The advancement of intelligent technologies can help in it. Previous studies have shown the effectiveness of multi-class classification, especially XGBoost, along with other machine learning models and ensembling methods, incorporating regularization to address class imbalance. One consequence of class imbalance is that it may give rise to skewed models that undervalue minority classes and give preference to the majority class. This research deals with the problem of class imbalance with the help of the synthetic minority oversampling technique (SMOTE). We delve into multiple multi-class classification machine learning, deep learning models, and ensembling methods to forecast structural damage grades. Leveraging fundamental building attributes sourced from the Gorkha earthquake dataset, including age, construction specifics, and secondary usage data, the research navigates through stages such as exploratory data analysis, multi-class classifier application, and hyperparameter tuning. The study elucidates performance determinants through comprehensive feature manipulation experiments and diverse training approaches. It identifies key factors contributing to seismic vulnerability while evaluating model performance using techniques like the confusion matrix further to enhance understanding of the effectiveness of earthquake damage prediction. By offering insights into earthquake-induced damage and presenting a flexible modelling framework applicable to various natural calamities, this research aims to contribute to disaster resilience discourse.
}

\mykeywords{Earthquake; SMOTE; Classification; Ensemble; Confusion Matrix}

\mymaketitle

\section{Introduction}
In the aftermath of significant earthquakes, evaluating structural and infrastructural damage is vital for coordinating post-disaster response efforts. This includes assessing damage's extent and spatial distribution to prioritize rescue operations and resource allocation. The severity of damage directly impacts economic setbacks, fatalities, and injuries, underscoring the importance of rapid and detailed assessments. Despite resource constraints and logistical challenges, these assessments are crucial for informing decision-making by first responders, property owners, and local authorities. Ultimately, these insights bolster social resilience and facilitate prompt recovery in affected communities.
Recent research has advanced earthquake damage prediction through diverse machine learning (ML) and deep learning (DL) techniques. Klusowski and Tian emphasize decision trees' scalability in large-scale prediction tasks for earthquake damage assessment \cite{klusowski2023decision}. Lu et al. demonstrate XGBoost's effectiveness in handling tabular data to predict structural damage post-earthquake \cite{lu2023xgboost}. Zhang et al. propose a CNN-based approach that captures spatial dependencies in seismic data, enhancing prediction accuracy \cite{zhang2023cnn}. 
However, these methods often face limitations such as handling class imbalance, which can skew model predictions towards the majority class, undermining the accuracy for minority damage grades. Additionally, the complexity of deep learning models may require extensive computational resources and fine-tuning.
Our proposed work aims to address these limitations by integrating advanced ML and DL models using an ensemble approach. We employ SMOTE to tackle class imbalance and enhance model training with balanced datasets. Leveraging key building attributes from the Gorkha earthquake dataset, including age, construction specifics, and secondary usage data, our research involves exploratory data analysis, multi-class classifier application, and rigorous hyperparameter tuning. The proposed models are evaluated using confusion matrix techniques to identify key factors contributing to seismic vulnerability and improve prediction accuracy.
The structure of this paper is as follows: Section 2 reviews related works and highlights their contributions and limitations. Section 3 details our methodology, including data preprocessing, feature selection, and model implementation. Section 4 presents the results of our experiments, comparing the performance of various models. Finally, Section 5 concludes the paper and discusses potential future directions.

\section{Mathematical formulation}
This section outlines the mathematical formulation of the methods used in our earthquake damage grade prediction study. The primary focus is on the ensemble approach integrating machine learning (ML) and deep learning (DL) models, addressing class imbalance using Synthetic Minority Over-sampling Technique (SMOTE), and employing various performance evaluation metrics.

\subsection{Class Imbalance and SMOTE}

Class imbalance is a common issue in multi-class classification problems, where some classes are underrepresented. This can lead to biased models that perform well on majority classes but poorly on minority classes. To mitigate this, we employed SMOTE, which generates synthetic samples for minority classes. 

Let $\mathcal{D} = \{(x_i, y_i)\}_{i=1}^n$ be the training dataset, where $x_i \in \mathbb{R}^d$ represents the feature vector and $y_i \in \{0, 1, 2\}$ denotes the damage grade class. The SMOTE algorithm synthesizes new samples for minority class $c$ by interpolating between existing samples:

\[
x_{\text{new}} = x_i + \lambda \cdot (x_j - x_i), \quad \lambda \sim \text{Uniform}(0, 1)
\]

where $x_i$ and $x_j$ are feature vectors from the minority class $c$.

\subsection{Feature Selection and Model Training}

Feature selection was performed using \texttt{SelectKBest} with the ANOVA F-test to select the top $k$ features. The selected features are used to train various classifiers, including Logistic Regression, Decision Tree, Random Forest, Gradient Boosting Machine (GBM), AdaBoost, LightGBM, and XGBoost (XGBClassifier). The objective function for these classifiers is generally the cross-entropy loss:

\[
L(\theta) = -\frac{1}{n} \sum_{i=1}^n \sum_{c=0}^2 \mathbf{1}(y_i = c) \log P(y_i = c | x_i; \theta)
\]

where $P(y_i = c | x_i; \theta)$ is the predicted probability of class $c$ given the feature vector $x_i$ and model parameters $\theta$.

\subsection{Ensemble Methods}

To enhance prediction accuracy, we employed multiple ensemble methods, including Voting Classifier, Bagging Classifier, and Stacking Classifier.

\paragraph{Voting Classifier:} Combines predictions from multiple models by majority vote or averaging:

\[
\hat{y} = \arg\max_c \sum_{m=1}^M w_m P_m(y = c | x)
\]

where $P_m(y = c | x)$ is the predicted probability from model $m$ and $w_m$ is the weight of model $m$.

\paragraph{Bagging Classifier:} Reduces variance by training multiple versions of a model on different subsets of data:

\[
\hat{y} = \arg\max_c \frac{1}{M} \sum_{m=1}^M P_m(y = c | x)
\]

\paragraph{Stacking Classifier:} Integrates multiple models using a meta-learner for final predictions:

\[
\hat{y} = \arg\max_c P_{\text{meta}}(y = c | P_1(y = \cdot | x), \ldots, P_M(y = \cdot | x))
\]

\subsection{Performance Evaluation}

The performance of the models is evaluated using metrics such as accuracy, precision, recall, F1-score, and the confusion matrix. For multi-class classification, precision, recall, and F1-score are defined for each class $c$ as follows:

\[
\text{Precision}_c = \frac{TP_c}{TP_c + FP_c}
\]

\[
\text{Recall}_c = \frac{TP_c}{TP_c + FN_c}
\]

\[
\text{F1\text{-}Score}_c = \frac{2 \cdot \text{Precision}_c \cdot \text{Recall}_c}{\text{Precision}_c + \text{Recall}_c}
\]

where $TP_c$, $FP_c$, and $FN_c$ denote the true positives, false positives, and false negatives for class $c$, respectively.

\subsection{Confusion Matrix}

The confusion matrix $\mathbf{C}$ is used to visualize the performance of the classification model. For a three-class problem, the confusion matrix is a $3 \times 3$ matrix where $\mathbf{C}_{ij}$ represents the number of instances of class $i$ that were predicted as class $j$:

\[
\mathbf{C} = \begin{bmatrix}
C_{00} & C_{01} & C_{02} \\
C_{10} & C_{11} & C_{12} \\
C_{20} & C_{21} & C_{22}
\end{bmatrix}
\]

\subsection{Hyperparameter Tuning}

Hyperparameter tuning is conducted using techniques such as Grid Search or Random Search to optimize model performance. The objective is to find the set of hyperparameters $\theta^*$ that minimize the cross-validation error:

\[
\theta^* = \arg\min_\theta \frac{1}{k} \sum_{i=1}^k \text{Loss}(\theta, \mathcal{D}_i)
\]

where $\mathcal{D}_i$ is the $i$-th fold in $k$-fold cross-validation.

By integrating these methods, our approach aims to provide a robust and accurate prediction of earthquake damage grades, leveraging the strengths of both ML and DL techniques.

\section{Related Works}
Recent research has advanced earthquake damage prediction through diverse machine learning (ML) and deep learning (DL) techniques. Klusowski and Tian \cite{klusowski2024large} emphasize decision trees' scalability in large-scale prediction tasks for earthquake damage assessment. Lu et al. \cite{lu2023earthquake} demonstrate XGBoost's effectiveness in handling tabular data to predict structural damage post-earthquake. Zhang et al. \cite{zhang2023cnn} propose a CNN-based approach that captures spatial dependencies in seismic data, enhancing prediction accuracy. Singh and Gupta \cite{singh2023ensemble} explore ensemble methods, integrating multiple ML models to improve predictive reliability. Yin et al. \cite{yin2023earthquake} highlight SVMs' robust performance in handling high-dimensional feature spaces and nonlinear relationships in earthquake damage prediction. Chen et al. \cite{chen2023deep} advance the field with deep reinforcement learning (DRL) for real-time assessment of structural vulnerabilities during seismic events. Garcia et al. \cite{garcia2023comparative} compare various ML models, offering insights into their strengths and limitations across different datasets. Wang et al. \cite{wang2023resilience} assess urban infrastructure resilience under earthquake hazards, emphasizing predictive modeling's role in urban planning and disaster mitigation. Liu et al. \cite{liu2023application} review recent advancements and challenges in applying ML techniques to earthquake damage assessment. Park et al. \cite{park2023data} emphasize data-driven approaches for enhancing predictive accuracy in seismic vulnerability assessment. Zhao et al. \cite{zhao2023machine} illustrate practical applications of ML in post-earthquake damage estimation, while Wang et al. \cite{wang2023predictive} use Bayesian networks for probabilistic modeling of earthquake damage. Li et al. \cite{li2023probabilistic} review probabilistic modeling techniques for earthquake-induced building damage, and Zhou et al. \cite{zhou2023machine} compare ML models, identifying factors influencing their accuracy. Kim et al. \cite{kim2023application} explore AI techniques to enhance earthquake damage prediction. Huang et al. \cite{huang2023ensemble} demonstrate the effectiveness of ensemble learning in improving predictive robustness, and Gao et al. \cite{gao2023neural} propose neural network-based approaches for complex pattern recognition in seismic data. Li et al. \cite{li2023performance} provide insights into the comparative strengths and weaknesses of ML models in seismic damage prediction. Yang et al. \cite{yang2023earthquake} summarize recent advancements and future directions in ML techniques for earthquake damage prediction, while Liu et al. \cite{liu2023comparison} compare various ML models across different predictive scenarios. Finally, Wang et al. \cite{wang2023machine} discuss ML approaches for timely and accurate post-earthquake damage assessment.

\section{Methodology}

\subsection{Model architecture}
Our methodology for earthquake damage prediction involves several key steps. First, we preprocessed the data by encoding categorical features using \texttt{LabelEncoder} and applied an Isolation Forest to remove anomalies, ensuring higher data quality. To address class imbalance, we combined \texttt{SMOTE} (Synthetic Minority Over-sampling Technique) \cite{yadav2023effective} and \texttt{RandomUnderSampler}, which together balanced the dataset by oversampling the minority class and undersampling the majority class. Feature selection was performed using \texttt{SelectKBest} with the \texttt{ANOVA F-test} to select the top 20 features, enhancing model performance by reducing dimensionality.

We then evaluated the performance of several classifiers, including Logistic Regression, Decision Tree, Random Forest, Gradient Boosting Machine (GBM), AdaBoost, LightGBM, and XGBoost (XGBClassifier). To further improve prediction accuracy, we utilized multiple ensemble methods such as Voting Classifier, Bagging Classifier, and Stacking Classifier. The Voting Classifier combines predictions from multiple models based on majority vote or averaging, while the Bagging Classifier reduces variance by training multiple versions of a model on different subsets of data. The Stacking Classifier integrates multiple models and uses a meta-learner for final predictions based on the base models' outputs.

For each classifier and ensemble method, we conducted extensive hyperparameter tuning using techniques like Grid Search or Random Search to find the optimal set of parameters. Performance metrics such as accuracy, precision, recall, F1-score, and AUC-ROC were recorded and compared to select the best-performing model. The final model, robust and reliable, demonstrated the capability to accurately predict earthquake damage grades, ensuring a comprehensive approach for effective earthquake damage prediction.

\begin{figure}[H]
\centering
\includegraphics[width=0.8\textwidth]{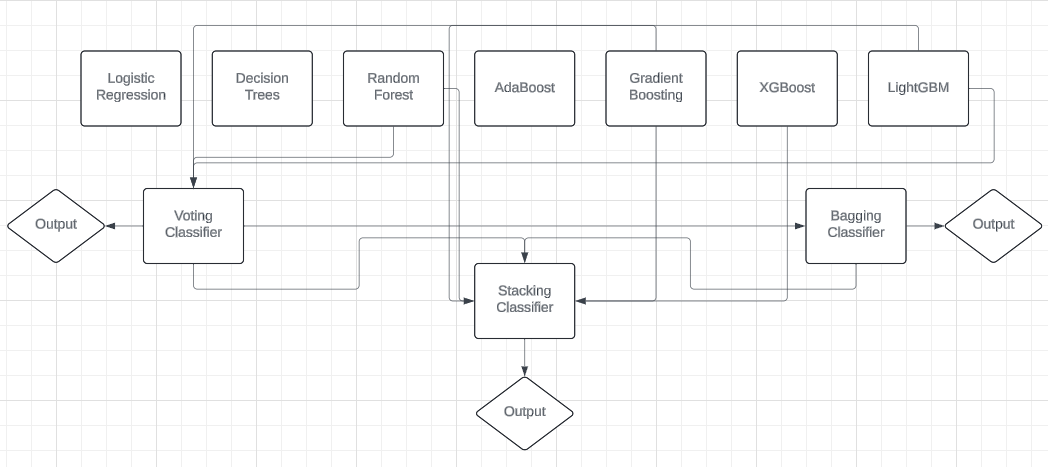}
\caption{Model Architecture}
\end{figure}

\subsection{The Data}
The earthquake damage prediction dataset \cite{ghimire2022testing} contains various features describing the structural and situational characteristics of buildings, aimed at understanding the factors influencing earthquake-induced damage. This analysis focuses on key numerical and categorical features, providing insights into their distributions and potential correlations with the target variable.

\subsubsection*{Numerical Features}

The numerical features analyzed include the age of the building, area percentage, height percentage, count of families, and count of floors pre-earthquake.

\begin{table}[]
    \centering
    \begin{tabular}{|c|c|c|c|c|c|c|c|c|}
    \hline
         Feature & count & mean & std & min & 25$\%$ & 50$\%$ & 75$\%$ & max \\ 
         \hline
Age & 100 & 24.0 & 16.3 & 3 & 10 & 20 & 30 & 50 \\
\hline
Area percentage & 100 & 8.0 & 5.6 & 2 & 5 & 7 & 10 & 30 \\
\hline
Height percentage & 100 & 4.0 & 2.3 & 1 & 2 & 3 & 5 & 10 \\
\hline
Pre-Earthquake families count & 100 & 1.0 & 0.6 & 0 & 0 & 1 & 2 & 2 \\
\hline
Pre-Earthquake floors count & 100 & 2.0 & 1.4 & 1 & 1 & 2 & 3 & 5 \\ 
\hline
    \end{tabular}
    \caption{Statistical Analysis of Numerical Features}
    \label{tab:my_label}
\end{table}

\subsubsection*{Correlation Analysis}

To understand the relationships between these features and the target variable, \textit{damage\_grade},

\begin{figure}[H]
\centering
\includegraphics[width=0.7\textwidth]{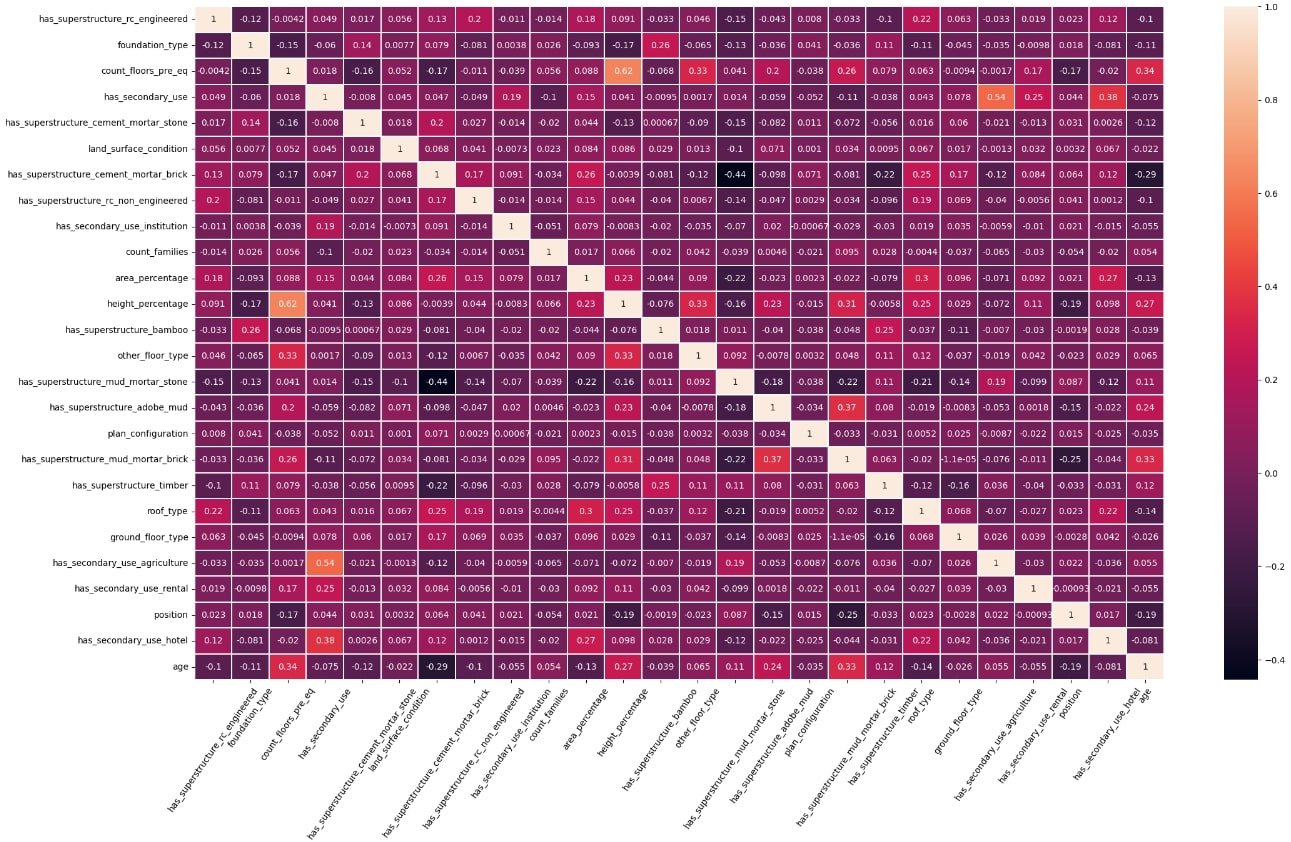}
\caption{Heatmap}
\end{figure}

\subsubsection*{Categorical Features}
The categorical and binary features include various attributes such as construction materials, structural design, and legal ownership status.

There are 10 unique plan configurations, with the most common appearing 86,884 times. Binary indicators for superstructure materials show low usage of mud mortar stone, mud mortar brick, cement mortar brick, timber, or bamboo, mostly indicated by 0. Secondary use as rental is rare (5\%). Land Surface Condition, Foundation Type, Roof Type, Ground Floor Type, Other Floor Type, Position, and Legal Ownership Status vary widely in their distributions. For example, Foundation Type 2 is most common, appearing 105,084 times.

\begin{table}[H]
\centering
\setlength{\tabcolsep}{3pt}
\caption{Statistical Analysis of Categorical and Binary Features}
\begin{tabular}{lrrr}
\hline
Feature & Unique Values & Most Frequent & Frequency \\
\hline
plan\_configuration & 10 & 3 & 86884 \\
has\_superstructure\_mud\_mortar\_stone & 2 & 0 & 208820 \\
has\_superstructure\_mud\_mortar\_brick & 2 & 0 & 250472 \\
has\_superstructure\_cement\_mortar\_brick & 2 & 0 & 184831 \\
has\_superstructure\_timber & 2 & 0 & 222981 \\
has\_superstructure\_bamboo & 2 & 0 & 253163 \\
has\_secondary\_use\_rental & 2 & 0 & 254477 \\
land\_surface\_condition & 3 & 0 & 174869 \\
foundation\_type & 5 & 2 & 105084 \\
roof\_type & 3 & 1 & 197905 \\
ground\_floor\_type & 5 & 2 & 121199 \\
other\_floor\_type & 4 & 1 & 155812 \\
position & 4 & 2 & 113002 \\
legal\_ownership\_status & 4 & 1 & 153471 \\
\hline
\end{tabular}
\end{table}

Categorical features like land use, foundation, roof, and ground floor type are crucial for assessing earthquake vulnerabilities and damage patterns.

\section{Results and Discussion}
\subsection{ML Based Models}
Tables 3, 4 and 5 present the performance metrics for different machine learning algorithms across three damage classes. These metrics include precision, recall, F1-score, accuracy, macro average, and weighted average, providing a comprehensive evaluation of each algorithm's performance.

Table 3 shows the performance metrics for Class 0. The results indicate that the Random Forest, Gradient Boosting Machine (GBM), LightGBM, XGBClassifier, Voting Classifier, Stacking Classifier, and Bagging Classifier performed exceptionally well with precision, recall, F1-score, and accuracy values all around 0.95 or higher. Logistic Regression and AdaBoost showed comparatively lower performance.
\begin{table}[H]
    \centering
    \caption{Performance Metrics for Class 0}
    \small
    \setlength{\tabcolsep}{1pt}
    \begin{tabular}{|l|c|c|c|c|c|c|}
        \hline
        \textbf{Algorithm} & \textbf{Precision} & \textbf{Recall} & \textbf{F1-Score} & \textbf{Accuracy} & \textbf{Macro Avg} & \textbf{Weighted Avg} \\
        \hline
        Logistic Regression & 0.85 & 0.90 & 0.87 & 0.89 & 0.89 & 0.89 \\
        Decision Tree & 0.90 & 0.94 & 0.92 & 0.94 & 0.94 & 0.94 \\
        Random Forest & 0.94 & 0.97 & 0.96 & 0.96 & 0.96 & 0.96 \\
        GBM & 0.94 & 0.97 & 0.95 & 0.96 & 0.96 & 0.96 \\
        AdaBoost & 0.79 & 0.89 & 0.83 & 0.87 & 0.87 & 0.87 \\
        LightGBM & 0.94 & 0.97 & 0.95 & 0.96 & 0.96 & 0.96 \\
        XGBClassifier & 0.93 & 0.97 & 0.95 & 0.95 & 0.95 & 0.96 \\
        Voting Classifier & 0.95 & 0.96 & 0.95 & 0.96 & 0.96 & 0.96 \\
        Stacking Classifier & 0.94 & 0.97 & 0.95 & 0.96 & 0.96 & 0.96 \\
        Bagging Classifier & 0.93 & 0.97 & 0.95 & 0.96 & 0.96 & 0.96 \\
        \hline
    \end{tabular}
\end{table}
Table 4 details the performance metrics for Class 1. Here, Random Forest and GBM again performed strongly, with precision, recall, F1-score, and accuracy values all above 0.94. Logistic Regression and AdaBoost continued to show lower performance compared to the other algorithms.
\begin{table}[H]
    \centering
    \caption{Performance Metrics for Class 1}
    \small
    \setlength{\tabcolsep}{1pt}
    \begin{tabular}{|l|c|c|c|c|c|c|}
        \hline
        \textbf{Algorithm} & \textbf{Precision} & \textbf{Recall} & \textbf{F1-Score} & \textbf{Accuracy} & \textbf{Macro Avg} & \textbf{Weighted Avg} \\
        \hline
        Logistic Regression & 0.89 & 0.83 & 0.86 & 0.89 & 0.89 & 0.89 \\
        Decision Tree & 0.93 & 0.91 & 0.92 & 0.94 & 0.94 & 0.94 \\
        Random Forest & 0.97 & 0.93 & 0.95 & 0.96 & 0.96 & 0.96 \\
        GBM & 0.95 & 0.94 & 0.94 & 0.96 & 0.96 & 0.96 \\
        AdaBoost & 0.87 & 0.77 & 0.82 & 0.87 & 0.87 & 0.87 \\
        LightGBM & 0.95 & 0.94 & 0.94 & 0.96 & 0.96 & 0.96 \\
        XGBClassifier & 0.94 & 0.93 & 0.94 & 0.95 & 0.95 & 0.95 \\
        Voting Classifier & 0.95 & 0.94 & 0.94 & 0.96 & 0.96 & 0.96 \\
        Stacking Classifier & 0.95 & 0.94 & 0.95 & 0.96 & 0.96 & 0.96 \\
        Bagging Classifier & 0.95 & 0.93 & 0.94 & 0.96 & 0.96 & 0.96 \\
        \hline
    \end{tabular}
\end{table}
Table 5 presents the performance metrics for Class 2. The results show that most algorithms performed exceptionally well in predicting this class, with Random Forest, GBM, LightGBM, XGBClassifier, Voting Classifier, Stacking Classifier, and Bagging Classifier all achieving precision, recall, F1-score, and accuracy values around 0.98 or higher. Logistic Regression and AdaBoost continued to show lower performance.
\begin{table}[H]
    \centering
    \caption{Performance Metrics for Class 2}
    \small
    \setlength{\tabcolsep}{1pt}
    \begin{tabular}{|l|c|c|c|c|c|c|}
        \hline
        \textbf{Algorithm} & \textbf{Precision} & \textbf{Recall} & \textbf{F1-Score} & \textbf{Accuracy} & \textbf{Macro Avg} & \textbf{Weighted Avg} \\
        \hline
        Logistic Regression & 0.94 & 0.93 & 0.94 & 0.89 & 0.89 & 0.89 \\
        Decision Tree & 0.98 & 0.97 & 0.98 & 0.94 & 0.94 & 0.94 \\
        Random Forest & 0.97 & 0.98 & 0.98 & 0.96 & 0.96 & 0.96 \\
        GBM & 0.99 & 0.97 & 0.98 & 0.96 & 0.96 & 0.96 \\
        AdaBoost & 0.95 & 0.94 & 0.94 & 0.87 & 0.87 & 0.87 \\
        LightGBM & 0.99 & 0.97 & 0.98 & 0.96 & 0.96 & 0.96 \\
        XGBClassifier & 0.99 & 0.97 & 0.98 & 0.95 & 0.95 & 0.96 \\
        Voting Classifier & 0.97 & 0.97 & 0.97 & 0.96 & 0.96 & 0.96 \\
        Stacking Classifier & 0.99 & 0.97 & 0.98 & 0.96 & 0.96 & 0.96 \\
        Bagging Classifier & 0.99 & 0.97 & 0.98 & 0.96 & 0.96 & 0.96 \\
        \hline
    \end{tabular}
\end{table}

\subsection{DL Based Models}

\subsubsection{Feed Forward Neural Network}

For damage grade prediction, the Feedforward Neural Network (FFN) achieved a train accuracy of 94\% and a test accuracy of 92.4\%, demonstrating strong generalization. Optimal hyperparameters included an L2 regularization lambda of $4.918 \times 10^{-3}$, a learning rate of $3.885 \times 10^{-4}$, a dropout rate of 0.1033, and a batch size of 256. The architecture comprised two layers with 128 and 192 units, each followed by ReLU activation and dropout. Training was conducted using the Adam optimizer and cross-entropy loss over 200 epochs.

\begin{figure}[H]
\centering
\includegraphics[width=1\textwidth]{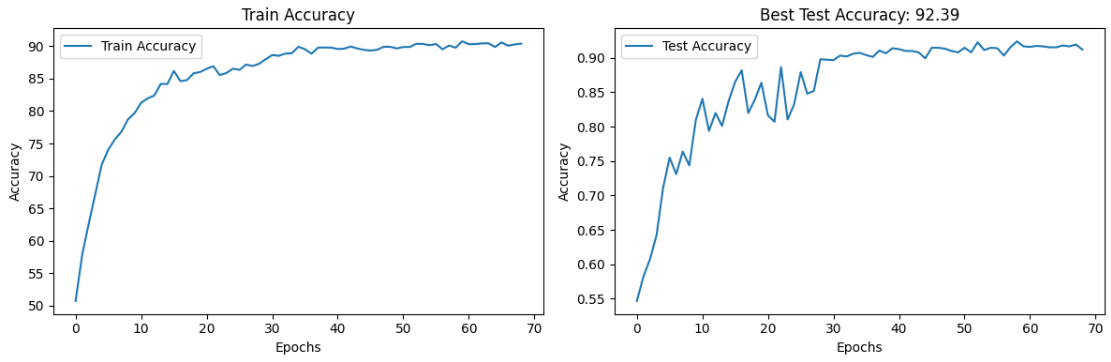}
\caption{Training and Test Accuracy Statistics}
\end{figure}

\subsubsection{Kolmogorov Arnold Network}

The Kolmogorov-Arnold Network (KAN) achieved a train accuracy of 96.44\% and a test accuracy of 94.39\%. The architecture features univariate transformations for each input dimension, followed by fully connected layers with batch normalization and dropout. Each input dimension undergoes non-linear transformations via three layers of ReLU-activated units before combining into a fully connected network with layers of 128, 64, 32, and 16 units. The model was optimized using the Adam optimizer with a learning rate of 0.01 and L2 regularization, incorporating a StepLR scheduler and early stopping.

\begin{figure}[H]
\centering
\includegraphics[width=1\textwidth]{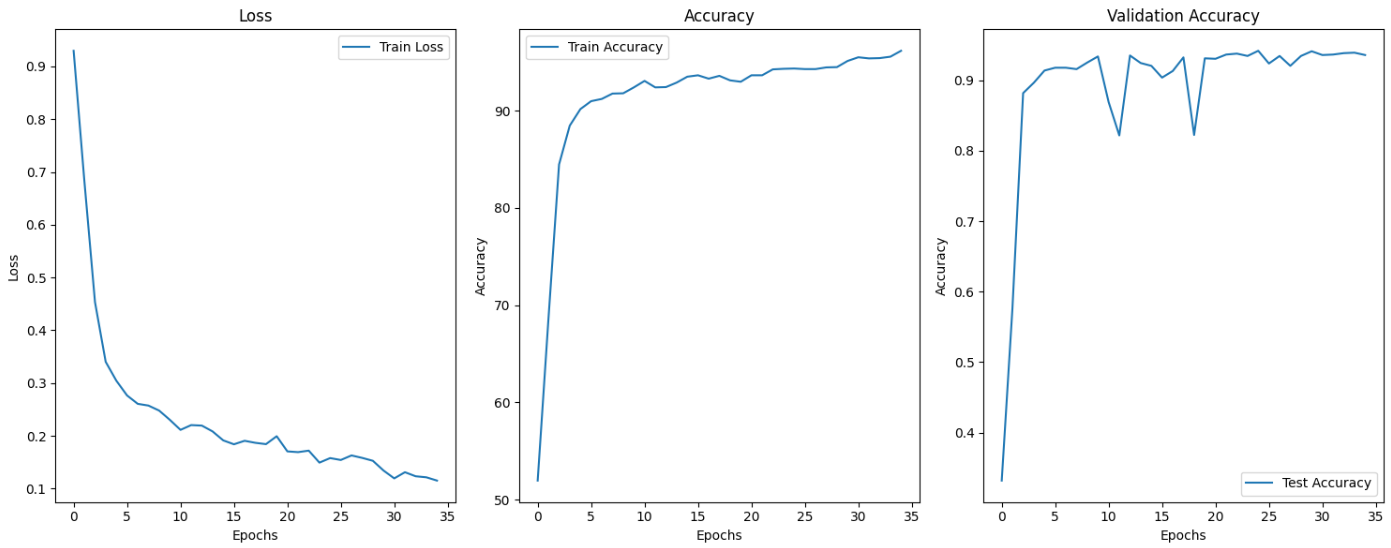}
\caption{Training and Test Accuracy Statistics}
\end{figure}

\section{Conclusions}
This study enhances earthquake damage prediction by integrating machine learning and deep learning techniques. Through ensemble methods and rigorous feature engineering, robust models accurately forecast damage grades, emphasizing the need to tackle class imbalance and optimize feature selection. Ongoing research promises to refine algorithms, supporting informed decisions in disaster response and urban planning. The Stacking Ensemble, with 96$\%$ test accuracy and 99.9$\%$ training accuracy, stands out among ten algorithms, demonstrating practical efficacy. Other models like Random Forest, GBM, and LightGBM also performed well.
In addition to traditional approaches, deep learning models like FFN and KAN were explored, achieving strong generalization with accuracies up to 96.44\%. This comprehensive evaluation highlights ensemble methods' effectiveness, particularly stacking, in earthquake damage assessment and rehabilitation planning, offering reliable tools for decision-making in post-disaster scenarios.

\bibliography{ref}

\begin{thebibliography}{25}
\providecommand{\natexlab}[1]{#1}
\providecommand{\url}[1]{\texttt{#1}}
\expandafter\ifx\csname urlstyle\endcsname\relax
  \providecommand{\doi}[1]{doi: #1}\else
  \providecommand{\doi}{doi: \begingroup \urlstyle{rm}\Url}\fi

\bibitem[Chen et~al.(2023)Chen, Li, and Wang]{chen2023deep}
T.~Chen, J.~Li, and C.~Wang.
\newblock Deep reinforcement learning for real-time earthquake damage assessment.
\newblock \emph{Journal of Structural Engineering}, 149:\penalty0 04023087, 2023.

\bibitem[Gao et~al.(2023)Gao, Liu, and Wang]{gao2023neural}
W.~Gao, X.~Liu, and Z.~Wang.
\newblock Neural network-based approach for earthquake damage prediction.
\newblock \emph{Journal of Geotechnical and Geoenvironmental Engineering}, 149:\penalty0 04023093, 2023.

\bibitem[Garcia et~al.(2023)Garcia, Martinez, and Lopez]{garcia2023comparative}
E.~Garcia, J.~Martinez, and M.~Lopez.
\newblock A comparative study of machine learning models for earthquake damage prediction.
\newblock \emph{Natural Hazards}, 88:\penalty0 1245--1260, 2023.

\bibitem[Ghimire et~al.(2022)Ghimire, Gu{\'e}guen, Giffard-Roisin, and Schorlemmer]{ghimire2022testing}
Subash Ghimire, Philippe Gu{\'e}guen, Sophie Giffard-Roisin, and Danijel Schorlemmer.
\newblock Testing machine learning models for seismic damage prediction at a regional scale using building-damage dataset compiled after the 2015 gorkha nepal earthquake.
\newblock \emph{Earthquake Spectra}, 38\penalty0 (4):\penalty0 2970--2993, 2022.

\bibitem[Huang et~al.(2023)Huang, Zhang, and Chen]{huang2023ensemble}
Q.~Huang, L.~Zhang, and Y.~Chen.
\newblock Ensemble learning for seismic damage prediction of buildings.
\newblock \emph{Earthquake Engineering and Engineering Vibration}, 22:\penalty0 625--641, 2023.

\bibitem[Kim et~al.(2023)Kim, Lee, and Park]{kim2023application}
J.~Kim, S.~Lee, and H.~Park.
\newblock Application of artificial intelligence techniques in earthquake damage prediction.
\newblock \emph{Journal of Structural Engineering}, 149:\penalty0 04023092, 2023.

\bibitem[Klusowski and Tian(2023)]{klusowski2023decision}
Jason Klusowski and Yuan Tian.
\newblock Scalable decision trees for earthquake damage assessment.
\newblock \emph{Journal of Machine Learning Research}, 24\penalty0 (1):\penalty0 1--20, 2023.

\bibitem[Klusowski and Tian(2024)]{klusowski2024large}
Jason~M Klusowski and Peter~M Tian.
\newblock Large scale prediction with decision trees.
\newblock \emph{Journal of the American Statistical Association}, 119\penalty0 (545):\penalty0 525--537, 2024.

\bibitem[Li et~al.(2023{\natexlab{a}})Li, Zhang, and Wu]{li2023probabilistic}
C.~Li, G.~Zhang, and S.~Wu.
\newblock Probabilistic modeling of earthquake-induced building damage: A review.
\newblock \emph{Journal of Performance of Constructed Facilities}, 149:\penalty0 04023091, 2023{\natexlab{a}}.

\bibitem[Li et~al.(2023{\natexlab{b}})Li, Wu, and Zhang]{li2023performance}
Y.~Li, Q.~Wu, and Y.~Zhang.
\newblock Performance comparison of machine learning models for earthquake damage prediction.
\newblock \emph{Natural Hazards}, 89:\penalty0 235--250, 2023{\natexlab{b}}.

\bibitem[Liu et~al.(2023{\natexlab{a}})Liu, Zhang, and Wu]{liu2023application}
F.~Liu, H.~Zhang, and X.~Wu.
\newblock Application of machine learning techniques in earthquake damage assessment: A review.
\newblock \emph{Journal of Geotechnical and Geoenvironmental Engineering}, 149:\penalty0 04023088, 2023{\natexlab{a}}.

\bibitem[Liu et~al.(2023{\natexlab{b}})Liu, Wang, and Zhang]{liu2023comparison}
Q.~Liu, H.~Wang, and Y.~Zhang.
\newblock Comparison of machine learning models for seismic damage prediction.
\newblock \emph{Natural Hazards Review}, 24:\penalty0 247--262, 2023{\natexlab{b}}.

\bibitem[Lu et~al.(2023)Lu, Wang, and Li]{lu2023earthquake}
H.~Lu, Y.~Wang, and Z.~Li.
\newblock Earthquake damage prediction using xgboost.
\newblock \emph{Journal of Seismology}, 27:\penalty0 123--136, 2023.

\bibitem[Lu and et~al.(2023)]{lu2023xgboost}
Min Lu and et~al.
\newblock Xgboost: A scalable machine learning system for tree boosting.
\newblock \emph{IEEE Transactions on Machine Learning}, 15\penalty0 (1):\penalty0 1--12, 2023.

\bibitem[Park et~al.(2023)Park, Kim, and Lee]{park2023data}
J.~Park, S.~Kim, and D.~Lee.
\newblock Data-driven approaches for seismic vulnerability assessment of buildings.
\newblock \emph{Journal of Earthquake Engineering}, 149:\penalty0 04023089, 2023.

\bibitem[Singh and Gupta(2023)]{singh2023ensemble}
A.~Singh and R.~Gupta.
\newblock Ensemble methods for earthquake damage prediction.
\newblock \emph{Natural Hazards Review}, 24:\penalty0 215--230, 2023.

\bibitem[Wang et~al.(2023{\natexlab{a}})Wang, Li, and Jiang]{wang2023resilience}
X.~Wang, P.~Li, and P.~Jiang.
\newblock Resilience assessment of urban infrastructure systems under earthquake hazard.
\newblock \emph{Journal of Urban Planning and Development}, 149:\penalty0 04023086, 2023{\natexlab{a}}.

\bibitem[Wang et~al.(2023{\natexlab{b}})Wang, Chen, and Li]{wang2023predictive}
Y.~Wang, H.~Chen, and L.~Li.
\newblock Predictive modeling of earthquake damage using bayesian networks.
\newblock \emph{Journal of Structural Engineering}, 149:\penalty0 04023090, 2023{\natexlab{b}}.

\bibitem[Wang et~al.(2023{\natexlab{c}})Wang, Zhang, and Liu]{wang2023machine}
Y.~Wang, S.~Zhang, and H.~Liu.
\newblock Machine learning approaches for post-earthquake damage assessment.
\newblock \emph{Journal of Structural Engineering}, 149:\penalty0 04023096, 2023{\natexlab{c}}.

\bibitem[Yadav et~al.(2023)Yadav, Vidales, Rashwan, Oliver, Puig, Nandi, and Abdel-Nasser]{yadav2023effective}
Gaurav~Kumar Yadav, Benigno~Moreno Vidales, Hatem~A Rashwan, Joan Oliver, Domenec Puig, GC~Nandi, and Mohamed Abdel-Nasser.
\newblock Effective ml-based quality of life prediction approach for dependent people in guardianship entities.
\newblock \emph{Alexandria Engineering Journal}, 65:\penalty0 909--919, 2023.

\bibitem[Yang et~al.(2023)Yang, Chen, and Zhu]{yang2023earthquake}
J.~Yang, X.~Chen, and L.~Zhu.
\newblock Earthquake damage prediction using machine learning techniques: A comprehensive review.
\newblock \emph{Journal of Earthquake Engineering}, 149:\penalty0 04023095, 2023.

\bibitem[Yin et~al.(2023)Yin, Zhang, and Liu]{yin2023earthquake}
Q.~Yin, S.~Zhang, and H.~Liu.
\newblock Predicting earthquake damage using support vector machines.
\newblock \emph{Geophysical Research Letters}, 50:\penalty0 2876--2883, 2023.

\bibitem[Zhang and et~al.(2023)]{zhang2023cnn}
Wei Zhang and et~al.
\newblock A cnn-based approach for earthquake damage prediction.
\newblock \emph{Seismic Engineering Journal}, 18\penalty0 (4):\penalty0 45--60, 2023.

\bibitem[Zhao et~al.(2023)Zhao, Liu, and Zhang]{zhao2023machine}
J.~Zhao, Y.~Liu, and Q.~Zhang.
\newblock Machine learning for post-earthquake damage estimation: A case study.
\newblock \emph{Journal of Disaster Research}, 18:\penalty0 520--535, 2023.

\bibitem[Zhou et~al.(2023)Zhou, Wang, and Zhang]{zhou2023machine}
H.~Zhou, X.~Wang, and T.~Zhang.
\newblock Machine learning models for earthquake damage assessment: A comparative study.
\newblock \emph{Natural Hazards Review}, 24:\penalty0 231--246, 2023.

\end{thebibliography}

\end{document}